\documentclass{sig-alternate-05-2015}
\usepackage{graphicx}
\usepackage{caption}
\usepackage{verbatim}
\usepackage{subcaption}
\usepackage{amsmath}
\graphicspath{ {figures/} }
\usepackage{array}
\newcolumntype{P}[1]{>{\centering\arraybackslash}p{#1}}
\newcolumntype{M}[1]{>{\centering\arraybackslash}m{#1}}
\usepackage{caption}
\begin{document}
\title{Joint Deep Modeling of Users and Items Using Reviews for Recommendation}
%
%
%
%
%

\numberofauthors{3} 
%
\author{
%
%
 \alignauthor
Lei Zheng\\
\affaddr{Department of Computer Science}\\
\affaddr{University of Illinois at Chicago}\\
\affaddr{Chicago, U.S.}\\
\email{lzheng21@uic.edu}
\alignauthor Vahid Noroozi\\
         \affaddr{Department of Computer Science}\\
        \affaddr{University of Illinois at Chicago}\\
        \affaddr{Chicago, U.S.}\\
        \email{vnoroo2@uic.edu}\\
 \alignauthor Philip S. Yu\\
         \affaddr{Department of Computer Science}\\
        \affaddr{University of Illinois at Chicago}\\
        \affaddr{Chicago, U.S.}\\
        \email{psyu@uic.edu}
 }
\CopyrightYear{2017} 
\setcopyright{acmcopyright}
\conferenceinfo{WSDM 2017,}{February 06-10, 2017, Cambridge, United Kingdom}
\isbn{978-1-4503-4675-7/17/02}\acmPrice{\$15.00}
\doi{http://dx.doi.org/10.1145/3018661.3018665}
\maketitle
\begin{abstract}
A large amount of information exists in reviews written by users. This source of information has been ignored by most of the current recommender systems while it can potentially alleviate the sparsity problem and improve the quality of recommendations. In this paper, we present a deep model to learn item properties and user behaviors jointly from review text. The proposed model, named Deep Cooperative Neural Networks (DeepCoNN), consists of two parallel neural networks coupled in the last layers. One of the networks focuses on learning user behaviors exploiting reviews written by the user, and the other one learns item properties from the reviews written for the item. A shared layer is introduced on the top to couple these two networks together. The shared layer enables latent factors learned for users and items to interact with each other in a manner similar to factorization machine techniques. Experimental results demonstrate that DeepCoNN significantly outperforms all baseline recommender systems on a variety of datasets.
%
\end{abstract}


\section{Introduction}
\label{sec:intro}
The variety and number of products and services provided by companies have increased dramatically during the last decade. Companies produce a large number of products to meet the needs of customers. Although this gives more options to customers, it makes it harder for them to process the large amount of information provided by companies. Recommender systems help customers by presenting products or services that are likely of interest to them based on their preferences, needs, and past buying behaviors. Nowadays, many people use recommender systems in their daily life such as online shopping, reading articles, and watching movies.

Many of the prominent approaches employed in recommender systems \cite{koren2009matrix} are based on Collaborative Filtering (CF) techniques. The basic idea of these techniques is that people who share similar preferences in the past tend to have similar choices in the future. Many of the most successful CF techniques are based on matrix factorization \cite{koren2009matrix}. They find common factors that can be the underlying reasons for the ratings given by users. For example, in a movie recommender system, these factors can be genre, actors, or director of movies that may affect the rating behavior of users. Matrix factorization techniques not only find these hidden factors, but also learn their importance for each user and how each item satisfies each factor. 

Although CF techniques have shown good performance for many applications, the sparsity problem is considered as one of their significant challenges \cite{koren2009matrix}. The sparsity problem arises when the number of items rated by users is insignificant to the total number of items. It happens in many real applications. It is not easy for CF techniques to recommend items with few ratings or to give recommendations to the users with few ratings.

One of the approaches employed to address this lack of data is using the information in review text \cite{ling2014ratings,mcauley2013hidden}. In many recommender systems, other than the numeric ratings, users can write reviews for the products. Users explain the reasons behind their ratings in text reviews. The reviews contain information which can be used to alleviate sparsity problem. One of the drawbacks of most current CF techniques is that they model users and items just based on the numeric ratings provided by users and ignore the abundant information existed in the review text. Recently, some studies \cite{mcauley2013hidden} \cite{ling2014ratings} have shown that using review text can improve the prediction accuracy of recommender systems, in particular for the items and users with few ratings \cite{wang2010latent}.  

In this paper, we propose a neural network (NN) based model, named Deep \textbf{Co}operative \textbf{N}eural \textbf{N}etworks (DeepCoNN), to model users and items jointly using review text for rating prediction problems. The proposed model learns hidden latent features for users and items jointly using two coupled neural networks such that the rating prediction accuracy is maximized. One of the networks models user behavior using the reviews written by the user, and the other network models item properties using the written reviews for the item. The learned latent features for user and item are used to predict the corresponding rating in a layer introduced on the top of both networks. This interaction layer is motivated by matrix factorization techniques \cite{koren2009matrix} to let latent factors of users and items interact with each other. 

To the best of our knowledge, DeepCoNN is the first deep model that represents both users and items in a joint manner using reviews. It makes the model scalable and also suitable for online learning scenarios where the model needs to get updated continuously with new data. Another key contribution is that DeepCoNN represents review text using pre-trained word-embedding technique \cite{mikolov2013distributed, mikolov2010recurrent} to extract semantic information from the reviews. Recently, this representation has shown excellent results in many Natural Language Processing (NLP) tasks \cite{collobert2011natural,bengio2006neural,mikolov2013distributed}. Moreover, a significant advantage of DeepCoNN compared to most other approaches \cite{mcauley2013hidden,ling2014ratings} which benefit from reviews is that it models users and items in a joint manner with respect to prediction accuracy. Most of the similar algorithms perform the modeling independently of the ratings. Therefore, there is no guarantee that the learned factors can be beneficial to the rating prediction.

The experiments on real-world datasets including \textit{Yelp}, \textit{Amazon} \cite{mcauley2015inferring}, and \textit{Beer} \cite{mcauley2012learning} show that DeepCoNN outperforms all the compared baselines in prediction accuracy. Also, the proposed algorithm increases the performance for users and items with fewer ratings more than the ones with a higher number of ratings. It shows that DeepCoNN alleviates the sparsity problem by leveraging review text.

Our contributions and also advantages of DeepCoNN can be summarized as follows:
\begin{itemize}
\item The proposed Deep Cooperative Neural Networks (DeepCoNN) jointly model user behaviors and item properties using text reviews. The extra shared layer at the top of two neural networks connects the two parallel networks such that user and item representations can interact with each other to predict ratings. To the best of our knowledge, DeepCoNN is the first one that jointly models both user and item from reviews using neural networks.

\item It represents review text as word-embeddings using pre-trained deep models. The experimental results demonstrate that the semantic meaning and sentimental attitudes of reviews in this representation can increase the accuracy of rating prediction. All competing techniques which are based on topic modeling \cite{WuE15,BaoFZ14,DiaoQWSJW14} use the traditional \textit{bag of words} techniques.

 \item It does not only alleviate the problem of sparsity by leveraging reviews, but also improves the overall performance of the system significantly. It outperforms state-of-the-art techniques \cite{mcauley2013hidden, wang2011collaborative,DBLP:conf/nips/SalakhutdinovM07,wang2015collaborative} in terms of prediction accuracy on all of the evaluated datasets including Yelp, 21 categories of Amazon, and Beer (see Section \ref{sec:exp}).
  \end{itemize}
  
The rest of the paper is organized as follows. In Section \ref{sec:method}, we describe DeepCoNN in detail. Experiments are presented in Section \ref{sec:exp} to analyze DeepCoNN and demonstrate its effectiveness compared to the state-of-the-art techniques for recommendation systems.  In Section \ref{sec:pre}, we give a short review of the works related to our study. Finally, conclusions are presented in Section \ref{sec:con}.

\section{Methodology}
\label{sec:method}
The proposed model, DeepCoNN, is described in detail in this section. DeepCoNN models user behaviors and item properties using reviews. It learns hidden latent factors for users and items by exploiting review text such that the learned factors can estimate the ratings given by users. It is done with a CNN based model consisting of two parallel neural networks, coupled to each other with a shared layer at the top. The networks are trained in a joint manner to predict the ratings with minimum prediction error. We first describe notations used throughout this paper and formulate the definition of our problem. Then, the architecture of DeepCoNN and the objective function to get optimized is explained. Finally, we describe how to train this model.
\subsection{Definition and Notation}
A set of training set $\mathcal{T}$ consists of $N$ tuples. Each tuple $(u, i, r_{ui}, w_{ui})$ denotes a review written by user $u$ for item $i$ with rating $r_{ui}$ and text review of $w_{ui}$. The mathematical notations used in this paper are summarized in Table \ref{notation}.
\begin{table}[h]
\small
\centering
\caption{\textbf{Notations}}
\label{notation}
\begin{tabular}{|c|c|}
\textbf{Symbols} & \textbf{Definitions and Descriptions} \\
\hline
$d^u_{1:n}$ &user or item $u$'s review text consisting of $n$\\
            &  words \\
$V^u_{1:n}$ & word vectors of user or item $u$\\
$w_{ui}$ & a review text written by user $u$ for item $i$\\
${ o }_{j}$ & the output of $j_{th}$ neuron in the convolutional\\
            & layer\\
$n_{i}$ & the number of neurons in the layer $i$ \\
$K_j$ & the $j_{th}$ kernel in the convolutional layer\\
${ b }_{ j }$ & the bias of $j_{th}$ convolutional kernel\\
$g$ & the bias of the fully connected layer\\
$z_j$ & the $j_{th}$ feature map in the convolutional layer\\
$W$ & the weight matrix of the fully connected layer\\
$t$ & the window size of convolutional kernel\\
$c$ & the dimension of word embedding\\
$\mathbf{x_u}$ & the output of $Net_u$\\
$\mathbf{y_i}$ & the output of $Net_i$\\
$\lambda$ & the learning rate\\
\end{tabular}
\end{table}%
\subsection{Architecture}
\begin{figure*}[t]
\centering
\includegraphics[width=.6\textwidth,height=.39\textheight]{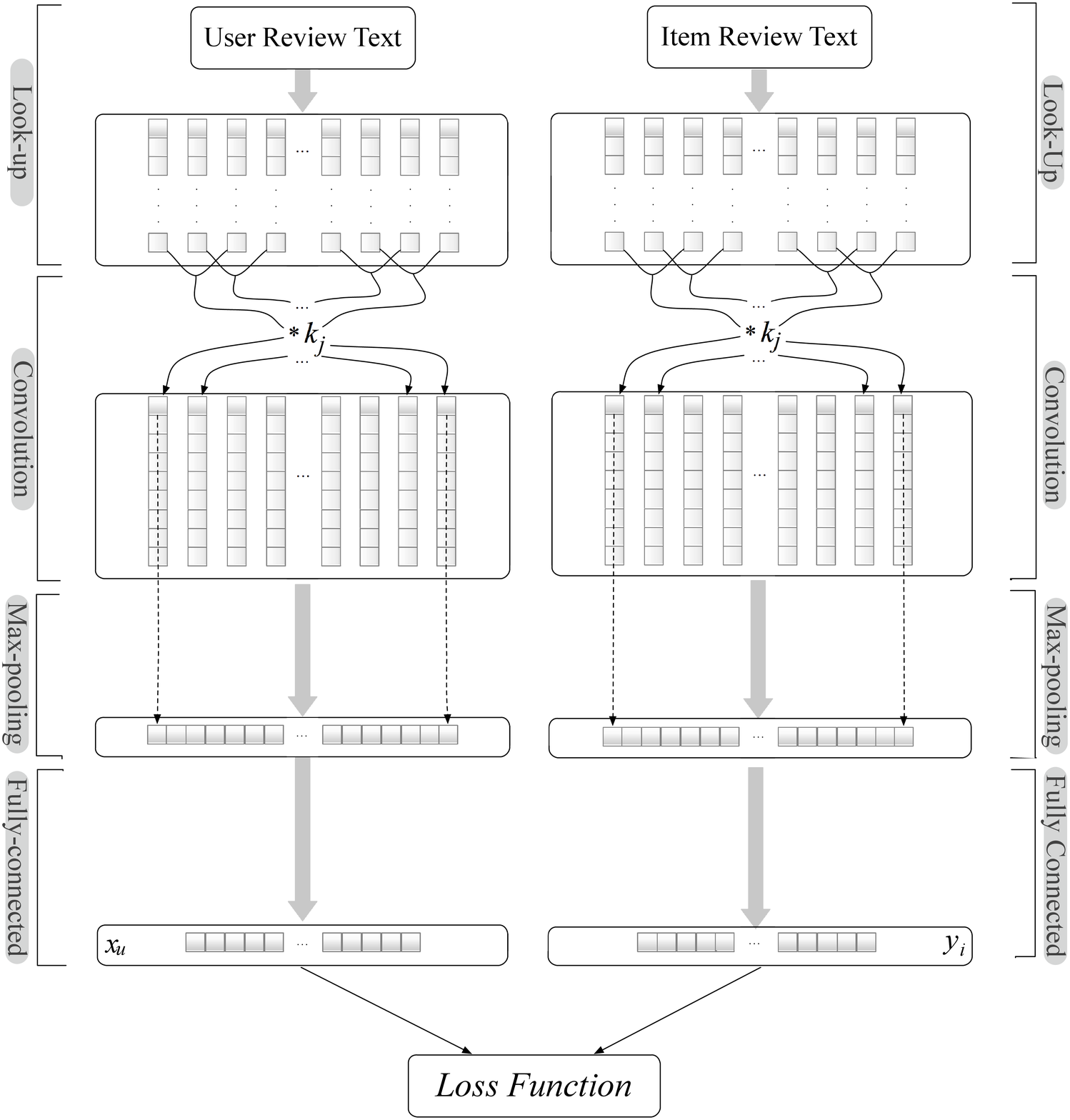}
\caption{The architecture of the proposed model}
\label{architecture}
\end{figure*}

The architecture of the proposed model for rating prediction is shown in Figure \ref{architecture}. The model consists of two parallel neural networks coupled in the last layer, one network for users ($Net_u$) and one network for items ($Net_i$). User reviews and item reviews are given to $Net_u$ and $Net_i$ respectively as inputs, and corresponding rating is produced as the output. In the first layer, denoted as look-up layer, review text for users or items are represented as matrices of word embeddings to capture the semantic information in the review text. Next layers are the common layers used in CNN based models to discover multiple levels of features for users and items, including convolution layer, max pooling layer, and fully connected layer. Also, a top layer is added on the top of the two networks to let the hidden latent factors of user and item interact with each other. This layer calculates an objective function that measures the rating prediction error using the latent factors produced by $Net_u$ and $Net_i$. In the following subsections, since $Net_u$ and $Net_i$ only differ in their inputs, we focus on illustrating the process for $Net_u$ in detail. The same process is applied for $Net_i$ with similar layers.
\subsection{Word Representation}
A word embedding $f:M \rightarrow \Re^n$, where $M$ represents the dictionary of words, is a parameterized function mapping words to $n$-dimensional distributed vectors. Recently, this approach has boosted the performance in many NLP applications \cite{kim2014convolutional, collobert2011natural}. 
DeepCoNN uses this representation technique to exploit the semantics of reviews. In the look-up layer, reviews are represented as a matrix of word embeddings to extract their semantic information. To achieve it, all the reviews written by user $u$, denoted as user reviews, are merged into a single document $d^u_{1:n}$, consisting of $n$ words in total. Then, a matrix of word vectors, denoted as $V^u_{1:n}$, is built for user $u$ as follows:
\begin{equation}
\label{eq1}
{ V }^u_{ 1:n }=\phi(d^u_{ 1 })\oplus \phi(d^u_{ 2 })\oplus \phi(d^u_{ 3 })\oplus \quad ...\quad \oplus \phi(d^u_{ n }),
\end{equation}
where $d^u_k$ indicates the $k$th word of document $d^u_{1:n}$, look-up function $\phi(d^u_k)$ returns the corresponding $c$-dimensional word vector for the word $d^u_k$, and $\oplus$ is the concatenation operator. It should be considered that the order of words is preserved in matrix $V^u_{1:n}$ that is another advantage of this representation comparing to \textit{bag-of-words} techniques.  

\subsection{CNN Layers}
Next layers including convolution layer, max pooling, and fully connected layer follow the CNN model introduced in \cite{collobert2011natural}. Convolution layer consists of $m$ neurons which produce new features by applying convolution operator on word vectors ${ V }^u_{ 1:n }$ of user $u$. Each neuron $j$ in the convolutional layer uses filter $K_j \in \Re^{c \times t}$ on a window of words with size $t$. For ${ V }^u_{ 1:n }$, we perform a convolution operation regarding each kernel $K_j$ in the convolutional layer.
\begin{equation}
\label{eq2}
z_j=f({ V }^u_{ 1:n }\ast K_j+b_j)
\end{equation}
Here symbol $\ast$ is convolution operator, $b_{j}$ is a bias term and $f$ is an activation function. In the proposed model, we use Rectified Linear Units (ReLUs) \cite{nair2010rectified}. It is defined as Eq. \ref{relu}. Deep convolutional neural networks with ReLUs train several times faster than their equivalents with tanh units \cite{krizhevsky2012imagenet}.
\begin{equation}
\label{relu}
f(x) = max\{0,x\}
\end{equation}
Following the work of \cite{collobert2011natural}, we then apply Eq. \ref{eq4}, a max pooling operation, over the feature map and take the maximum value as the feature corresponding to this particular kernel. The most important feature of each feature map, which has the highest value, has been captured. This pooling scheme can naturally deal with the varied length of the text. After the max pooling operation, convolutional results are reduced to a fixed size vector. 
\begin{equation}
\label{eq4}
o_j=max\{ z_1,z_2,...,z_{(n-t+1)}\} 
\end{equation}
We have described the process by which one feature is extracted from one kernel. The model uses multiple filters to obtain various features and the output vector of the convolutional layer is as Eq. \ref{cv_out}. 
\begin{equation}
\label{cv_out}
O=\{o_{1},o_{2},o_{3},...,o_{n_1}\},
\end{equation}
where $n_1$ denotes the number of kernel in the convolutional layer.
\begin{equation}
\label{fully}
\mathbf{x_u}=f(W \times O+g)
\end{equation}

The results from the max-pooling layer are passed to a fully connected layer with weight matrix $W$. As shown in Eq. \ref{fully}, the output of the fully connected layer $\mathbf{x_u} \in \Re^{n_2 \times 1}$  is considered as features for user $u$. Finally, the outputs of both user and item CNN $\mathbf{x_u}$ and $\mathbf{y_i}$ can be obtained.

\subsection{The Shared Layer}
Although these outputs can be viewed as features of users and items, they can be in different feature space and not comparable. Thus, to map them into the same feature space, we introduce a shared layer on the top to couple $Net_u$ and $Net_i$. First, let us concatenate $\mathbf{x_u}$ and $\mathbf{y_i}$ into a single vector $\mathbf{\hat{z}} = (\mathbf{x_u},\mathbf{y_i})$. To model all nested variable interactions in $\mathbf{\hat{z}}$, we introduce Factorization Machine (FM) \cite{rendle2012factorization} as the estimator of the corresponding rating. Therefore, given a batch of $N$ training examples $\mathcal{T}$, we can write down its cost as Eq. \ref{loss}.
\begin{equation}
\label{loss}
J={ \hat { w }  }_{ 0 }+\sum _{ i=1 }^{ |\hat{z}| }{ { \hat{w} }_{ i }{ \hat{z}  }_{ i }+\sum _{ i=1 }^{ |\hat{z} | }{ \sum _{ j=i+1 }^{ |\hat{z} | }{ \left< { \hat{\mathbf{v}} }_{ i },{ \hat{\mathbf{v}} }_{ j } \right>  } { \hat{z}}_{ i }{\hat{z} }_{ j } }  },
\end{equation}
where $\hat {{ w }  }_{ 0 }$ is the global bias, ${ \hat{w} }_{ i }$ models the strength of the $i_{th}$ variable in $\hat{z}$ and $\left< { \hat{\mathbf{v}} }_{ i },{ \hat{\mathbf{v}} }_{ j } \right> = \sum _{ f=1 }^{ |\hat{z} | }{\hat{\mathbf{v}}_{ i,f }\hat{\mathbf{v}}_{ j,f }}$. $\left< { \hat{\mathbf{v}} }_{ i },{ \hat{\mathbf{v}} }_{ j } \right>$ models the second order interactions. 
\subsection{Network Training}
Our network is trained by minimizing Eq. \ref{loss}. We take derivatives of $J$ with respect to $z$, as shown in Eq. \ref{deri}.
\begin{equation}
\label{deri}
\frac { \partial J }{ \partial { \hat{z} }_{ i } } =\hat{w}_{ i } + \sum _{ j=i+1 }^{ |\hat{z} | }{ \left< { \hat{\mathbf{v}} }_{ i },{ \hat{\mathbf{v}} }_{ j } \right>  }{\hat{z} }_{ j }   
\end{equation}

The derivatives of other parameters in different layers can be computed by applying differentiation chain rule. 

Given a set of training set $\mathcal{T}$ consisting of $N$ tuples, we optimize the model through RMSprop \cite{tieleman2012lecture} over shuffled mini-batches. RMSprop is an adaptive version of gradient descent which adaptively controls the step size with respect to the absolute value of the gradient. It does it by scaling the update value of each weight by a running average of its gradient norm. The updating rules for parameter set $\theta$ of the networks are as the following:
 \begin{equation}
 \label{rmsprop1}
 r_t \leftarrow 0.9{(\frac{\partial J}{ \partial \theta  })}^2 + 0.1r_{t-1}
 \end{equation}
 \begin{equation}
 \label{rmsprop2}
 \theta \leftarrow \theta - (\frac{\lambda}{\sqrt[]{r_t}+\epsilon}) \frac{\partial J}{ \partial \theta  },
 \end{equation}
 where $\lambda$ is the learning rate, $\epsilon$ is a small value added for numerical stability. Additionally, to prevent overfitting, the dropout \cite{srivastava2014dropout} strategy has also been applied to the fully connected layers of the two networks. 
\subsection{Some Analysis on DeepCoNN}
\subsubsection{Word Order Preservation}
Most of the recommender systems which use reviews in the modeling process employ topic modeling techniques to model users or items \cite{chen2015recommender}. Topic modeling techniques infer latent topic variables using the \textit{bag-of-words} assumption, in which word order is ignored. However, in many text modeling applications, word order is crucial \cite{wallach2006topic}. DeepCoNN is not based on topic modeling and uses word embeddings to create a matrix of word vectors where the order of words are preserved. In this way, convolution operations make use of the internal structure of data and provide a mechanism for efficient use of words' order in text modeling \cite{DBLP:conf/naacl/Johnson015}.  
\subsubsection{Online Learning}
Scalability and handling dynamic pools of items and users are considered as critical needs of many recommender systems. The time sensitivity of recommender systems poses a challenge in learning latent factors in an online fashion. DeepCoNN is scalable to the size of the training data, and also it can easily get trained and updated with new data because it is based on NN. Updating latent factors of items or users can get performed independently from historical data. All the approaches which employ topic modeling techniques do not benefit from these advantages to this extent. 

\section{Experiments}
\label{sec:exp}
We have performed extensive experiments on a variety of datasets to demonstrate the effectiveness of DeepCoNN compared to other state-of-the-art recommender systems. We first present the datasets and the evaluation metric used in our experiments in Section \ref{subsec:dataset}. The baseline algorithms selected for comparisons are explained in Section \ref{subsec:base}. Experimental settings are given in Section \ref{subsec:es}. Performance evaluation and some analysis of the model are discussed in sections \ref{subsec:pe} and \ref{sebsec:ma} respectively.

\subsection{Datasets and Evaluation Metric}
\label{subsec:dataset}
\begin{table*}[t]
\centering
\caption{\textbf{The Statistics of the datasets}}
\label{data}
\begin{tabular}{|c|c|c|c|c|c|c|}
\hline
Class & \#users & \#items & \#review & \#words & \#reviews per user&\#words per review \\ \hline
Yelp & 366,715 & 60,785 & 1,569,264 & 198M & 4.3 &126.41\\ \hline
Amazon & 6,643,669 & 2,441,053 & 34,686,770 & 4.053B & 5.2& 116.67   \\ \hline
Beer & 40,213 & 110,419 & 2,924,127 & 154M & 72.7& 52.67   \\ \hline
\end{tabular}
\label{dataset}
\end{table*}%
In our experiments, we have selected the following three datasets to evaluate our model.
\begin{itemize}
   \item[\textbullet] \textbf{Yelp}: It is a large-scale dataset consisting of restaurant reviews, introduced in the 6th round of \textit{Yelp Challenge} \footnote{\noindent https://www.yelp.com/dataset-challenge} in 2015. It contains more than 1M reviews and ratings. 
   
   \item[\textbullet] \textbf{Amazon}: Amazon Review dataset \cite{mcauley2015inferring} contains product reviews and metadata from Amazon website\footnote{\noindent https://snap.stanford.edu/data/web-Amazon.html}. It includes more than 143.7 million reviews spanning from May 1996 to July 2014. It has 21 categories of items, and as far as we know, this is the largest public available rating dataset with text reviews. 

   \item[\textbullet] \textbf{Beer}: It is a beer review dataset extracted from \textit{ratebeer.com}. The data span a period of more than 10 years, including almost 3 million reviews up to November 2011 \cite{mcauley2012learning}.
 \end{itemize}
 
As we can see in Table \ref{data}, all datasets contain more than half a million of reviews. However, in \textit{Yelp} and \textit{Amazon}, customers provide less than six pair of reviews and ratings on average which shows these two datasets are extremely sparse. This sparsity can largely deteriorate the performance of recommender systems. Besides, in all datasets, each review consists of less than 150 words on average. 

In our experiments, we adopt the well-known Mean Square Error (MSE) to evaluate the performance of the algorithms. It is selected because most of the related works have used the same evaluation metric\cite{mcauley2013hidden,ling2014ratings,almahairi2015learning}. MSE can be defined as follows:
\begin{equation}
\label{mse}
MSE=\frac { 1 }{ N } \sum _{ n=1 }^{ N }{ { ({ r }_{ n }-{ \hat { r }  }_{ n }) }^{ 2 } },
\end{equation}
where ${ r }_{ n }$ is the $n{\text{th}}$ observed value, ${ \hat { r }  }_{ n }$ is the $n{\text{th}}$ predicted value and $N$ is the total number of observations. 
\subsection{Baselines}
\label{subsec:base}
To validate the effectiveness of DeepCoNN, we have selected three categories of algorithms for evaluations: (i) purely rating based models. We chose Matrix Factorization (MF) and Probabilistic Matrix Factorization (PMF) to validate that review information is helpful for recommender systems, (ii) topic modeling based models which use review information. Most of the recommender systems which take reviews into consideration are based on topic modeling techniques. To compare our model with topic modeling based recommender systems, we select three representative models: Latent Dirichlet Allocation (LDA) \cite{blei2003latent}, Collaborative Topic Regression (CTR) \cite{wang2011collaborative} and Hidden Factor as Topic (HFT) \cite{mcauley2013hidden}, and (iii) deep recommender systems. In \cite{wang2015collaborative}, authors have proposed a state-of-the-art deep recommender system named Collaborative Deep Learning (CDL). Note that all the baselines except MF and PMF have incorporated review information into their models to improve prediction.
 \begin{itemize}
   \item[\textbullet] \textbf{MF}: \textbf{M}atrix \textbf{F}actorization \cite{koren2009matrix} is the most popular CF-based recommendation method. It only uses rating matrix as input and estimates two low-rank matrices to predict ratings. In our implementation, Alternating Least Squares (ALS) technique is adopted to minimize its objective function. 
   \item[\textbullet] \textbf{PMF}: \textbf{P}robabilistic \textbf{M}atrix \textbf{F}actorization is introduced in \cite{DBLP:conf/nips/SalakhutdinovM07}. It models latent factors of users and items by Gaussian distributions.
   \item[\textbullet] \textbf{LDA}: \textbf{L}atent \textbf{D}irichlet \textbf{A}llocation is a well-known topic modeling algorithm presented in \cite{blei2003latent}. In \cite{mcauley2013hidden}, it is proposed to employ LDA to learn a topic distribution from a set of reviews for each item. By treating the learned topic distributions as latent features for each item, latent features for each user is estimated by optimizing rating prediction accuracy with gradient descent. 
    \item[\textbullet] \textbf{CTR}: \textbf{C}ollaborative \textbf{T}opic \textbf{R}egression has been proposed by \cite{wang2011collaborative}. It showed very good performance on recommending articles in a one-class collaborative filtering problem where a user is either interested or not.
    \item[\textbullet] \textbf{HFT}: \textbf{H}idden \textbf{F}actor as \textbf{T}opic proposed in \cite{mcauley2013hidden} employs topic distributions to learn latent factors from user or item reviews. The authors have shown that item specific topic distributions produce more accurate predictions than user specific ones. Thus, we report the results of HFT learning from item reviews. 
    \item[\textbullet] \textbf{CDL}: \textbf{C}ollaborative \textbf{D}eep \textbf{L}earning tightly couples a Bayesian formulation of the stacked denoising auto-encoders and PMF. The middle layer of auto-encoders serves as a bridge between auto-encoders and PMF.
 \end{itemize}
\subsection{Experimental Settings}
\label{subsec:es}
\begin{figure}[t]
\centering
\includegraphics[width=.22\textwidth,height=.15\textheight]{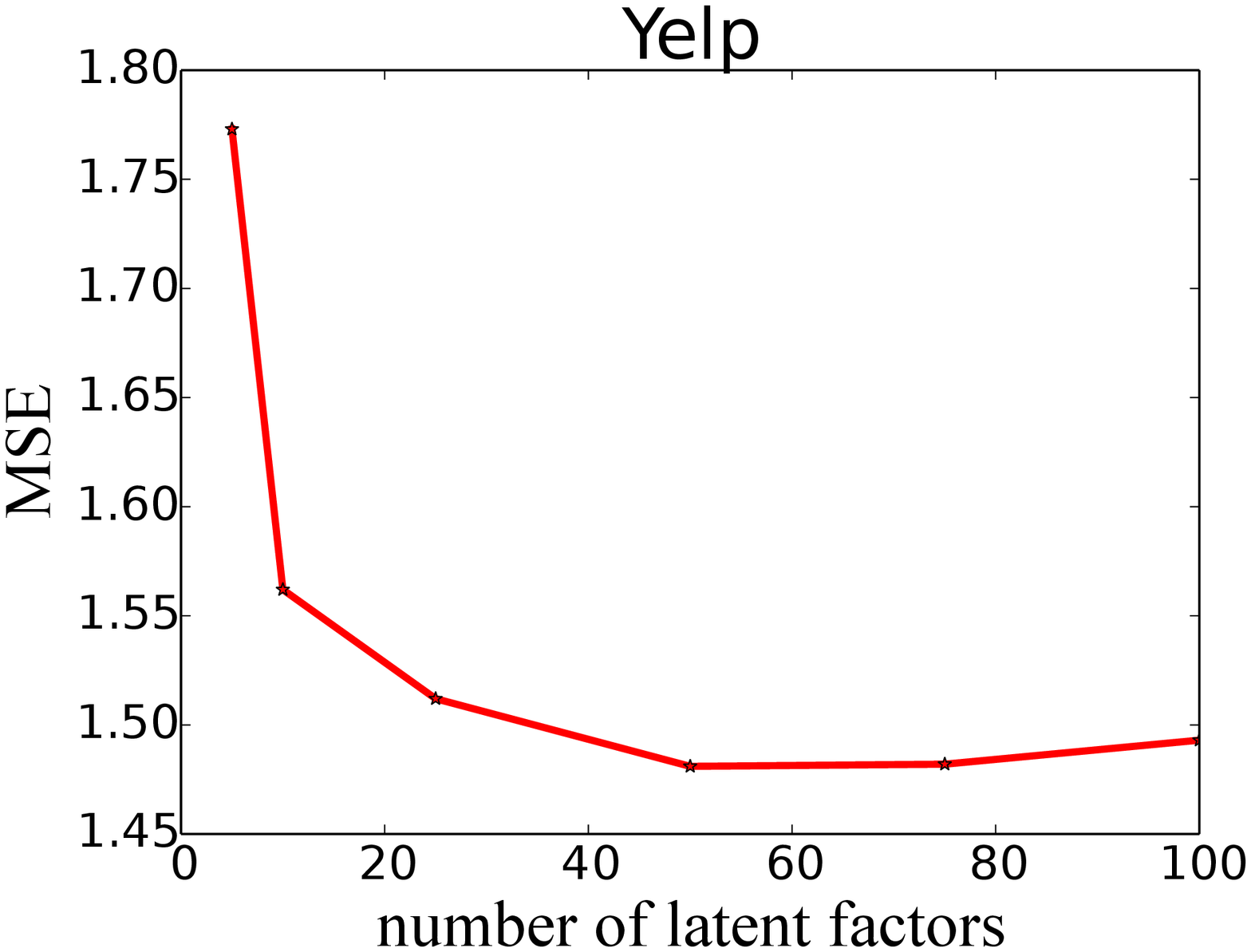}
\includegraphics[width=.22\textwidth,height=.15\textheight]{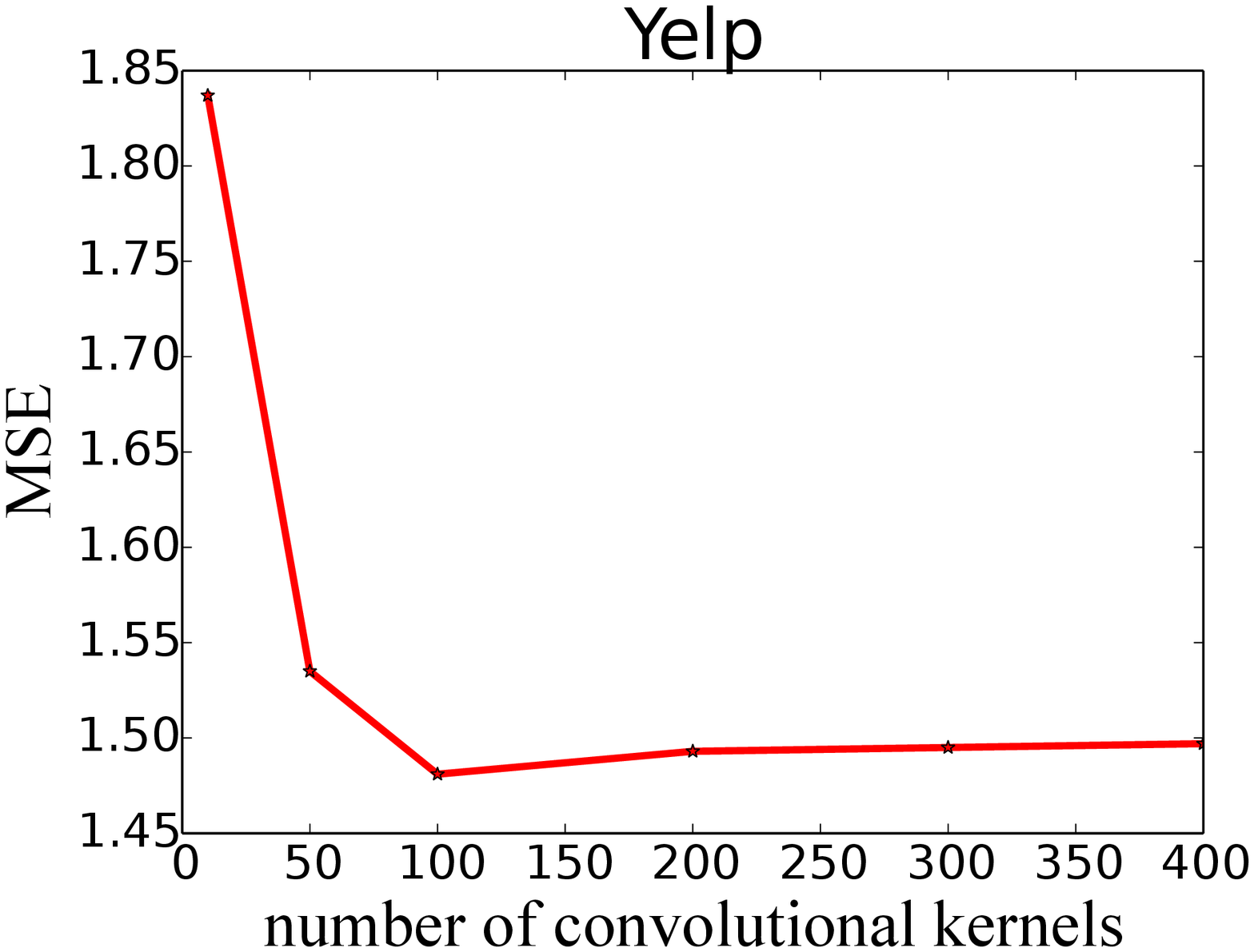}
\caption{The impact of the number of latent factors and
convolutional kernels on the performance of DeepCoNN in terms of MSE (\textit{Yelp} Dataset).}
\label{para_fig}
\end{figure}
We divided each dataset shown in Table \ref{data} into three sets of training set, validation set, and test set. We use 80\% of each dataset as the training set, 10\% is treated as the validation set to tune the hyper-parameters, and the rest is used as the test set. All the hyper-parameters of the baselines and DeepCoNN are selected based on the performance on the validation set.
 
For MF and PMF, we used grid search to find the best values for the number of latent factors from $\{25, 50, 100, 150, 200\}$, and regularization parameter from $\{0.001, 0.01, 0.1, 1.0\}$. 

For LDA, CTR and HFT, the number of topics $K$ is selected from \{5, 10, 20, 50, 100\} using the validation set. We set $K=10$ for LDA and CTR. The CTR model solves the \textit{one-class collaborative filtering problem} \cite{pan2008one} by using two different values for the precision parameter \textit{c} of a Gaussian distribution. Following the work of \cite{ling2014ratings}, in our experiments, we set precision \textit{c} as the same for all the observed ratings for rating prediction. HFT-k ($k=10,50$) are included to show the impact of the number of latent factors for HFT. By performing a grid search on the validation set, we set hyper-parameters $\alpha=0.1$, $\lambda_u=0.02$ and $\lambda_v=10$ for CTR and HFT. To optimize the performance of CDL, we performed a grid search on the hyper-parameters $\lambda_u$, $\lambda_v$, $\lambda_n$, $\lambda_w$ and $L$. Similar with CTR, the confidence parameter $c_{ij}$ of CDL is set as the same for all observed ratings.

We empirically studied the effects of two important parameters of DeepCoNN: the number of latent factors($|\mathbf{x_u}|$ and $|\mathbf{y_i}|$) and the number of convolutional kernels: $n_1$. In Figure \ref{para_fig}, we show the performance of DeepCoNN on the validation set of \textit{Yelp} with varying $|\mathbf{x_u}|$ and $|\mathbf{y_i}|$ from 5 to 100 and $n_1$ from 10 to 400 to investigate its sensitivity. As it can be seen, it does not improve the performance when the number of latent factors and number of kernels is greater than 50 and 100 respectively. Thus, we set $|\mathbf{x_u}|=|\mathbf{y_i}|=50$ and $n_1=100$. Other hyper-parameters: $t$, $c$, $\lambda$ and batch size are set as 3, 300, 0.002 and 100, respectively. These values were chosen through a grid search on the validation sets. We used a pre-trained word embeddings which are trained on more than 100 billion words from Google News \cite{mikolov2013distributed} \footnote{\noindent https://code.google.com/archive/p/word2vec/}.

Our models are implemented in \textit{Theano} \cite{2016arXiv160502688short}, a well-known Python library for machine learning and deep learning. The NVIDIA CUDA Deep Neural Network4 (cuDNN v4) accelerated our training process. All models are trained and tested on an NVIDIA Tesla K40 GPU. 
 \subsection{Performance Evaluation}
 \label{subsec:pe}
\begin{table*}[t]
\caption{\textbf{MSE Comparison with baselines}. \textbf{Best results are indicated in bold}.}
\centering
\label{res_3datasets}
\begin{tabular}{|c || c|c|c|c|c|c|c|c|M{2cm}|}
\hline
\textbf{Dataset} & MF& PMF &LDA&CTR&HFT-10&HFT-50&CDL&DeepCoNN&Improvement of DeepCoNN (\%) \\
\hline
Yelp & 1.792 & 1.783 & 1.788 & 1.612 & 1.583 & 1.587 & 1.574& \textbf{1.441} & 8.5\% \\
\hline
Amazon & 1.471 & 1.460 & 1.459 & 1.418 & 1.378 & 1.383 & 1.372& \textbf{1.268} & 7.6\% \\
\hline
Beer & 0.612 & 0.527 & 0.306 & 0.305 & 0.303 & 0.302 &  0.299&\textbf{0.273} & 8.7\% \\
\hline
\textbf{Average on all datasets} & 1.292 & 1.256 & 1.184 & 1.112 & 1.088 & 1.09 &  1.081&\textbf{0.994} & 8.3\% \\
\hline
\end{tabular}
\end{table*}%
The performance of DeepCoNN and the baselines (see Section \ref{subsec:base}) are reported in terms of MSE in Tables \ref{res_3datasets}. Table \ref{res_3datasets} shows the results on the three datasets including the performance averaged on all 21 categories of \textit{Amazon}. The experiments are repeated 3 times, and the averages are reported with the best performance shown in bold. The last column indicates the percentage of improvements gained by DeepCoNN compared to the best baseline in the corresponding category.

In Table \ref{res_3datasets}, all models perform better on \textit{Beer} dataset than on \textit{Yelp} and \textit{Amazon}. It is mainly related to the sparsity of \textit{Yelp} and \textit{Amazon}. Although PMF performs better than MF on \textit{Yelp}, \textit{Beer}, and most categories of \textit{Amazon}, both techniques do not show good performance compared to the ones which use reviews. It validates our hypothesis that review text provides additional information, and considering reviews in models can improve rating prediction. 

Although simply employing LDA to learn features from item reviews can help the model to achieve improvements, LDA models reviews independent of ratings. Therefore, there is no guarantee that the learned features can be beneficial to rating prediction. Therefore, by modeling ratings and reviews together, CTR and HFT attain additional improvements. Among those topic modeling based models (LDA, CTR and HFT), both HFT-10 and HFT-50 perform better in all three datasets. 

With the capability of extracting deep effective features from item review text, as we can see in Table \ref{res_3datasets}, CDL outperforms all topic modeling based recommender systems and advances the state-of-the-art. However, in benefiting from joint modeling capacity and semantic meaning existing from review text, DeepCoNN beats the best baseline in \textit{Yelp}, \textit{Beer} and \textit{Amazon} and gains \textbf{8.3\%} improvement on average.
\subsection{Model Analysis}
\begin{table}[t]
\centering
\caption{\textbf{Comparing variants of the proposed model}. \textbf{Best results are indicated in bold}.}
\label{res_var}
\begin{tabular}{|c || c|M{1.5cm}|c|}
\hline
\textbf{Model} & Yelp& Amazon Music Instruments&Beer\\
\hline
DeepCoNN-User & 1.577 & 1.373& 0.292\\ 
\hline
DeepCoNN-Item & 1.578 & 1.372  & 0.296\\ 
\hline
DeepCoNN-TFIDF & 1.713 &  1.469 & 0.589\\ 
\hline
DeepCoNN-Random &  1.799& 1.517  & 0.627\\ 
\hline
DeepCoNN-DP & 1.491 &  1.253 & 0.278 \\ 
\hline
DeepCoNN & \textbf{1.441} &  \textbf{1.233} & \textbf{0.273} \\ 
\hline
\end{tabular}
\end{table}%
\label{sebsec:ma}
Are the two parallel networks really cooperate to learn effective features from reviews? Does the proposed model benefit from the use of word embedding to exploit the semantic information in the review text? How much does the shared layer help in improving the predcition accuracy comparing to a simpler coupling approach? To answer these questions, we compare the DeepCoNN with its five variants: DeepCoNN-User, DeepCoNN-Item, DeepCoNN-TFIDF, DeepCoNN-Ra\\ndom and DeepCoNN-DP. These five variants are summarized in the following:
\begin{itemize}
 \item[\textbullet] \textbf{DeepCoNN-User}: The $Net_i$ of DeepCoNN is substituted with a matrix. Each row of the matrix is the latent factors of one item. This matrix is randomly initialized and optimized during the training.
 \item[\textbullet] \textbf{DeepCoNN-Item}: Similar with DeepCoNN-User, the $Net_u$ of DeepCoNN is replaced with a matrix. Each row of the matrix is the latent factors of one user. This matrix is randomly initialized and optimized during the training.
   \item[\textbullet] \textbf{DeepCoNN-TFIDF}: Instead of using word embedding, the TFIDF scheme is employed to represent review text as input to DeepCoNN.
    \item[\textbullet] \textbf{DeepCoNN-Random}: Our baseline model where all word representations are randomly initialized as fixed-length vectors. 
    \item[\textbullet] \textbf{DeepCoNN-DP}: The factorization machine in the objective function is substitued with a simple dot product of $\mathbf{x_u}$ and $\mathbf{y_i}$.
\end{itemize}

The performance of DeepCoNN and its variants on \textit{Yelp}, \textit{Beer} and one category of the \textit{Amazon} dataset: \textit{Music Instruments} are given in Table \ref{res_var}.

To demonstrate that the two deep CNNs can cooperate with each other to learn effective latent factors from user and item reviews, DeepCoNN-User and DeepCoNN-Item are trained with only one CNN with review text as input and the other CNN is substituted with a list of latent variables as the parameters to get learned. In this manner, latent factors of users or items are learned without considering their corresponding review text. As it can be seen in Table \ref{res_var}, while DeepCoNN-User and DeepCoNN-Item achieve similar results, DeepCoNN delivers the best performance by modeling both users and items. It verifies that review text is necessary for modeling latent factors of both users and items. Also, it shows that review text has informative information that can help to improve the performance of recommendation.

Furthermore, to validate the effectiveness of word representation, we compare DeepCoNN with DeepCoNN-TFIDF and DeepCoNN-Random. The DeepCoNN-TFIDF and Dee-\\pCoNN-Random are trained to show that word embedding is helpful to capture semantic meaning existed in the review text. While the performance of DeepCoNN-TFIDF is slightly better than  DeepCoNN-Random, they both perform considerably weaker than DeepCoNN. It shows the effectiveness of representing review text in semantic space for modeling the latent factors of items or users.

At last, to investigate the efficiency of the shared layer, DeepCoNN-DP is introduced that couples the two networks with a simpler objective function. The comparison shows the superiority of the factorization machine coupling. It can be the result of not only modeling the first order interactions but also the second order interactions between $\mathbf{x_u}$ and $\mathbf{y_i}$. 

\subsection{The Impact of the Number of Reviews}
\begin{figure*}[t]
\centering
\includegraphics[width=.3\textwidth,height=.18\textheight]{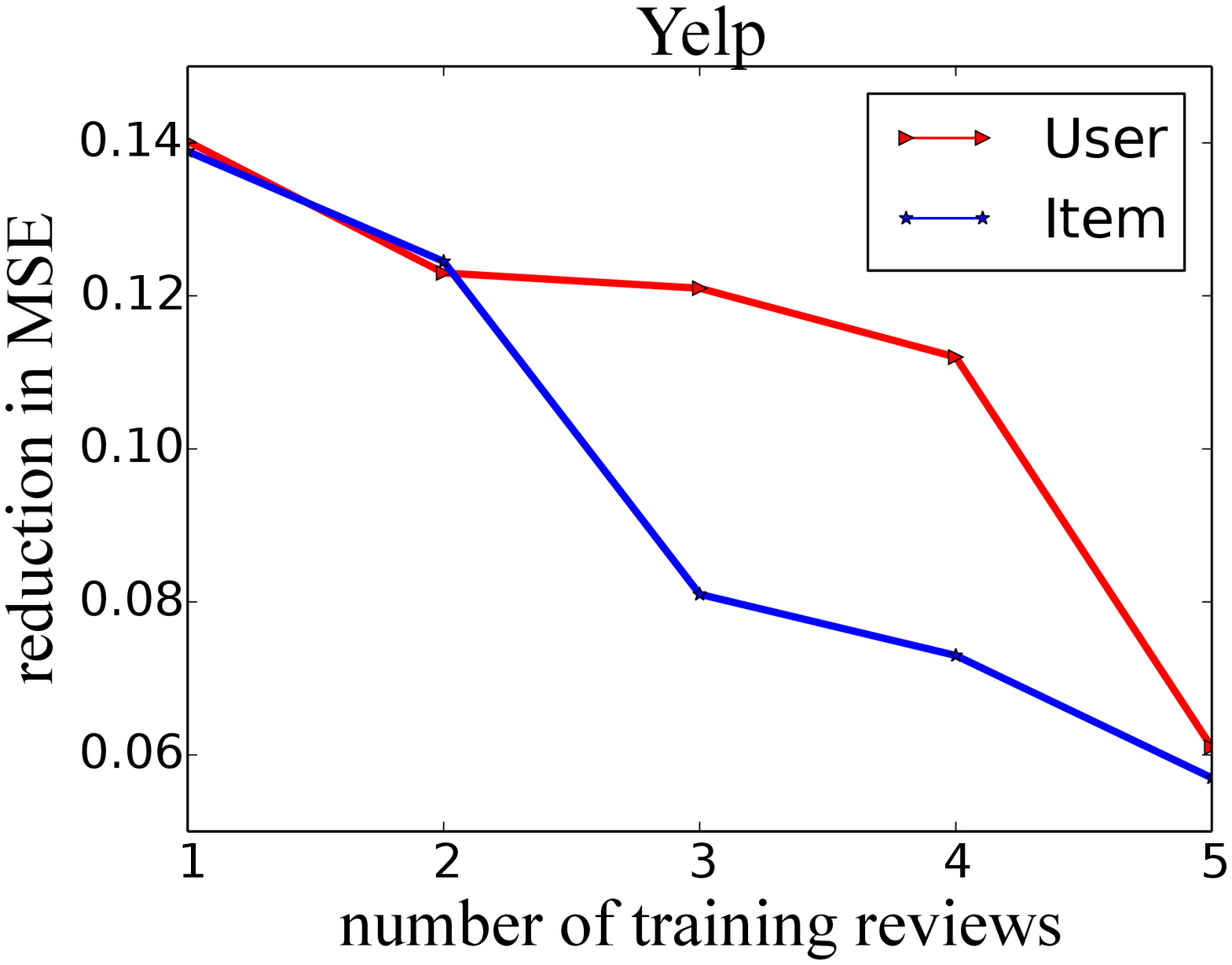}
\includegraphics[width=.3\textwidth,height=.18\textheight]{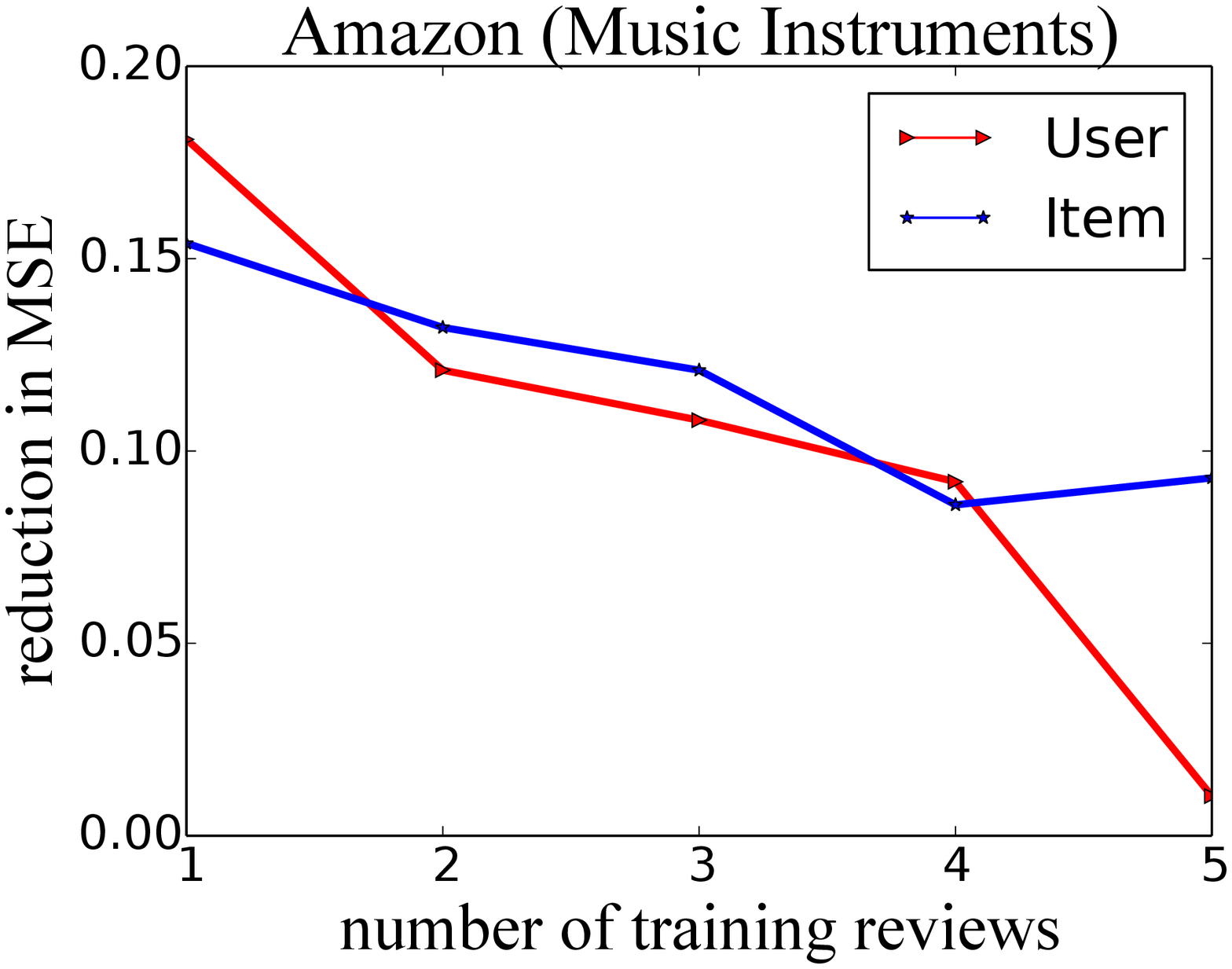}
\includegraphics[width=.3\textwidth,height=.18\textheight]{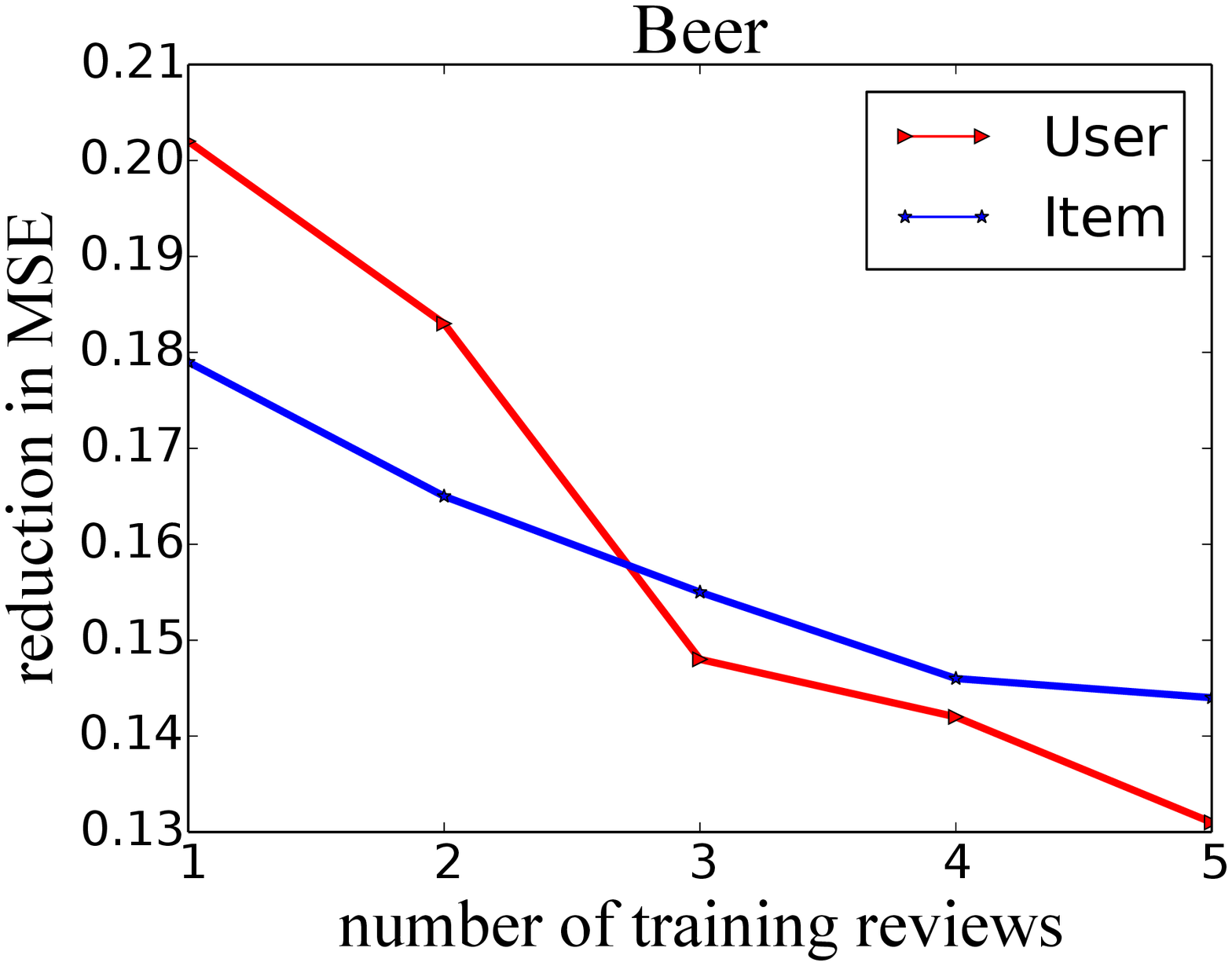}
\caption{MSE improvement achieved by DeepCoNN compared to MF. For users and items with different number of training reviews, DeepCoNN gains different MSE reductions.}
\label{num_review}
\end{figure*}
The cold start problem \cite{schein2002methods} is prevalent in recommender systems. In particular, when a new user joins or a new item is added to the system, their available ratings are limited. It would not be easy for the system to learn preferences of such users just from their ratings. It has been shown in some of the previous works that exploiting review text can help to alleviate this problem especially for users or items with few ratings \cite{mcauley2013hidden}. In this section, we conduct a set of experiments to answer the following questions. Can DeepCoNN help to tackle the cold start problem? What is the impact of the number of reviews on the effectiveness of the proposed algorithm? 

In Fig. \ref{num_review}, we have illustrated the reductions in MSE resulted from DeepCoNN compared to MF technique on three datasets of Yelp, Beer, and a group of Amazon (Music Instruments). By reduction in MSE, we mean the difference between the MSE of MF and the MSE of DeepCoNN. Users and items are categorized based on the number of their reviews, and reductions are plotted for both users and items groups. It can be seen that in all three datasets, reductions are positive, and DeepCoNN can achieve RMS reduction on all groups of users and items with few number of ratings. A more important advantage of DeepCoNN is that higher reductions are gained for groups with fewer ratings. It shows that DeepCoNN can alleviate the sparsity problem and help on the cold start problem.

It can also be seen that there exists a relation between the effectiveness of DeepCoNN and the number of ratings for a user or item. For users or items with a lower number of ratings, DeepCoNN reduction in MSE is higher. It shows that review text can be valuable information especially when we have limited information on the users or items.

\section{Related Works}
\label{sec:pre}
There are two categories of studies related to our work: techniques that model users and/or items by exploiting the information in online review text, and deep learning techniques employed for recommender systems. In this section, we give a short review of these two research areas and distinguish our work from the existing approaches.

The first studies that used online review text in rating prediction tasks were mostly focused on predicting ratings for an existing review \cite{baccianella2009multi,WuE15}, while in our paper, we predict the ratings from the history of review text written by a user to recommend desirable products to that user.


One of the pioneer works that explored using reviews to improve the rating prediction is presented in \cite{jakob2009beyond}. It found that reviews are usually related to different aspects, e.g., price, service, positive or negative feelings, that can be exploited for rating prediction. In \cite{mcauley2013hidden}, the authors proposed Hidden Factors as Topics (HFT) to employ topic modeling techniques to discover latent aspects from either item or user reviews. This method achieves significant improvement compared to models which only use ratings or reviews. A similar approach is followed in \cite{BaoFZ14} with the main difference that it models user's and items' reviews simultaneously. In \cite{DiaoQWSJW14}, a probabilistic model is proposed based on collaborative filtering and topic modeling. It uncovers aspects and sentiments of users and items, but it does not incorporate ratings during modeling reviews. Ratings Meet Reviews (RMR) \cite{ling2014ratings} also tries to harness the information of both ratings and reviews. One difference between HFT and RMR is that RMR applies topic modeling techniques on item review text and aligns the topics with the rating dimensions to improve prediction accuracy. 


Overall, one limitation of the above studies is that their textual similarity is solely based on lexical similarity. The vocabulary in English is very diverse, and two reviews can be semantically similar even with low lexical overlapping. The semantic meaning is of particular importance and has been ignored in these works. Additionally, reviews are represented by using \textit{bag-of-words}, and words' order exists in reviews has not been preserved. At last, the approaches which employ topic modeling techniques suffer from a scalability problem and also cannot deal with new coming users and items.

Recently, several studies have been done to use neural network based models including deep learning techniques for recommendation tasks. Several works \cite{salakhutdinov2007restricted,wucollaborative,li2015deep} model users and/or items from the rating matrix using neural networks like denoising auto-encoders or Restricted Boltzmann Machines (RBM). They are considered as collaborative based techniques because they just utilize the rating matrix and ignore review text unlike our approach.


In \cite{van2013deep} and \cite{wang2014improving}, deep models of CNN and Deep Belief Network (DBN) are introduced to learn latent factors from music data for music recommendation. In both models, initially, they find user and item latent factors using matrix factorization techniques. Then, they train a deep model such that it can reconstruct these latent factors for the items from the music content. A similar approach is followed in \cite{wang2015collaborative} for movie recommendation by using a generalized Stacked Auto Encoder (SAE) model. In all these works \cite{van2013deep, wang2014improving, wang2015collaborative}, an item's latent factors are learned from item's content and review text is ignored.

In \cite{elkahky2015multi}, a multi-view deep model is built to learn the user and item latent factors in a joint manner and map them to a common space. The general architecture of the model seems to have some similarities to our proposed model, but it differs from ours in some aspects. Their model is a content-based recommender system and does not use review text. Moreover, their outputs are coupled with a cosine similarity objective function to produce latent factors with high similarity. In this way, user and item factors are not learned explicitly in relation to the rating information, and there is no guarantee that the learned factors can help the recommendation task.

All the above NN based approaches differ from DeepCoNN because they ignore review text. To the best of our knowledge, the only work which has utilized deep learning techniques to use review text to improve recommendation is presented in \cite{almahairi2015learning}. To use the information exists in reviews, they proposed a model consisting of a matrix factorization technique and a Recurrent Neural Network (RNN). The matrix factorization is responsible for learning the latent factors of users and items, and the RNN models the likelihood of a review using the item's latent factors. The RNN model is combined with the MF simply via a trade-off term as some sort of a regularization term to tame the curse of data sparsity. Due to the matrix factorization technique, handling new users and items is not trivial in this model unlike DeepCoNN that handles them easily. Their proposed algorithm does not model users and items explicitly in a joint manner from their reviews, and it just uses reviews to regularize their model. In addition, since item text is represented by using \textit{bag-of-words}, semantic meaning existing in words has not been explored.  
\section{Conclusion}
\label{sec:con}
It is shown that reviews written by users can reveal some info on the customer buying and rating behavior, and also reviews written for items may contain info on their features and properties. In this paper, we presented Deep Cooperative Neural Networks (DeepCoNN) which exploits the information exists in the reviews for recommender systems. DeepCoNN consists of two deep neural networks coupled together by a shared common layer to model users and items from the reviews. It makes the user and item representations mapped into a common feature space. Similar to MF techniques, user and item latent factors can effectively interact with each other to predict the corresponding rating.

In comparison with state-of-the-art baselines, DeepCoNN achieved $\textbf{8.5\%}$ and $\textbf{7.6\%}$ improvements on datasets of $\textit{Yelp}$ and $\textit{Beer}$, respectively. On $\textit{Amazon}$, it outperformed all the baselines and gained $\textbf{8.7\%}$ improvement on average. Overall, $\textbf{8.3\%}$ improvement is attained by the proposed model on all three datasets. 

Additionally, in the experiments by limiting modeling to just one of the users and items, we demonstrated that the two networks could not only separately learn user and item latent factors from review text but also cooperate with each other to boost the performance of rating prediction. Furthermore, we showed that word embedding could be helpful to capture semantic meaning of review text by comparing it with a variant of DeepCoNN which uses random or TF-IDF representations for reviews.

At last, we conducted experiments to investigate the impact of the number of reviews. Experimental results showed that for the users and items with few reviews or ratings, DeepCoNN obtains more reduction in MSE than MF. Especially, when only one review is available, DeepCoNN gains the greatest MSE reduction. Thus, it validates that DeepCoNN can eﬀectively alleviate the sparsity problem.
\section{Acknowledgements}
This work is supported in part by NSF through grants IIS-1526499, and CNS-1626432. We gratefully acknowledge the support of NVIDIA Corporation with the donation of the Titan X GPU used for this research.
\bibliographystyle{abbrv}
\bibliography{bib}  

\begin{thebibliography}{10}

\bibitem{almahairi2015learning}
A.~Almahairi, K.~Kastner, K.~Cho, and A.~Courville.
\newblock Learning distributed representations from reviews for collaborative
  filtering.
\newblock In {\em Proceedings of the 9th ACM Conference on Recommender
  Systems}, pages 147--154. ACM, 2015.

\bibitem{baccianella2009multi}
S.~Baccianella, A.~Esuli, and F.~Sebastiani.
\newblock Multi-facet rating of product reviews.
\newblock In {\em Advances in Information Retrieval}, pages 461--472. Springer,
  2009.

\bibitem{BaoFZ14}
Y.~Bao, H.~Fang, and J.~Zhang.
\newblock Topicmf: Simultaneously exploiting ratings and reviews for
  recommendation.
\newblock In {\em {AAAI}}, pages 2--8. {AAAI} Press, 2014.

\bibitem{bengio2006neural}
Y.~Bengio, H.~Schwenk, J.-S. Sen{\'e}cal, F.~Morin, and J.-L. Gauvain.
\newblock Neural probabilistic language models.
\newblock In {\em Innovations in Machine Learning}, pages 137--186. Springer,
  2006.

\bibitem{blei2003latent}
D.~M. Blei, A.~Y. Ng, and M.~I. Jordan.
\newblock Latent dirichlet allocation.
\newblock {\em the Journal of machine Learning research}, 3:993--1022, 2003.

\bibitem{chen2015recommender}
L.~Chen, G.~Chen, and F.~Wang.
\newblock Recommender systems based on user reviews: the state of the art.
\newblock {\em User Modeling and User-Adapted Interaction}, 25(2):99--154,
  2015.

\bibitem{collobert2011natural}
R.~Collobert, J.~Weston, L.~Bottou, M.~Karlen, K.~Kavukcuoglu, and P.~Kuksa.
\newblock Natural language processing (almost) from scratch.
\newblock {\em The Journal of Machine Learning Research}, 12:2493--2537, 2011.

\bibitem{DiaoQWSJW14}
Q.~Diao, M.~Qiu, C.~Wu, A.~J. Smola, J.~Jiang, and C.~Wang.
\newblock Jointly modeling aspects, ratings and sentiments for movie
  recommendation {(JMARS)}.
\newblock In {\em {KDD}}, pages 193--202. {ACM}, 2014.

\bibitem{elkahky2015multi}
A.~M. Elkahky, Y.~Song, and X.~He.
\newblock A multi-view deep learning approach for cross domain user modeling in
  recommendation systems.
\newblock In {\em Proceedings of the 24th International Conference on World
  Wide Web}, pages 278--288. International World Wide Web Conferences Steering
  Committee, 2015.

\bibitem{jakob2009beyond}
N.~Jakob, S.~H. Weber, M.~C. M{\"u}ller, and I.~Gurevych.
\newblock Beyond the stars: exploiting free-text user reviews to improve the
  accuracy of movie recommendations.
\newblock In {\em Proceedings of the 1st international CIKM workshop on
  Topic-sentiment analysis for mass opinion}, pages 57--64. ACM, 2009.

\bibitem{DBLP:conf/naacl/Johnson015}
R.~Johnson and T.~Zhang.
\newblock Effective use of word order for text categorization with
  convolutional neural networks.
\newblock In {\em {HLT-NAACL}}, pages 103--112. The Association for
  Computational Linguistics, 2015.

\bibitem{kim2014convolutional}
Y.~Kim.
\newblock Convolutional neural networks for sentence classification.
\newblock {\em arXiv preprint arXiv:1408.5882}, 2014.

\bibitem{koren2009matrix}
Y.~Koren, R.~Bell, and C.~Volinsky.
\newblock Matrix factorization techniques for recommender systems.
\newblock {\em Computer}, (8):30--37, 2009.

\bibitem{krizhevsky2012imagenet}
A.~Krizhevsky, I.~Sutskever, and G.~E. Hinton.
\newblock Imagenet classification with deep convolutional neural networks.
\newblock In {\em Advances in neural information processing systems}, pages
  1097--1105, 2012.

\bibitem{li2015deep}
S.~Li, J.~Kawale, and Y.~Fu.
\newblock Deep collaborative filtering via marginalized denoising auto-encoder.
\newblock In {\em Proceedings of the 24th ACM International on Conference on
  Information and Knowledge Management}, pages 811--820. ACM, 2015.

\bibitem{ling2014ratings}
G.~Ling, M.~R. Lyu, and I.~King.
\newblock Ratings meet reviews, a combined approach to recommend.
\newblock In {\em Proceedings of the 8th ACM Conference on Recommender
  systems}, pages 105--112. ACM, 2014.

\bibitem{mcauley2013hidden}
J.~McAuley and J.~Leskovec.
\newblock Hidden factors and hidden topics: understanding rating dimensions
  with review text.
\newblock In {\em Proceedings of the 7th ACM conference on Recommender
  systems}, pages 165--172. ACM, 2013.

\bibitem{mcauley2012learning}
J.~McAuley, J.~Leskovec, and D.~Jurafsky.
\newblock Learning attitudes and attributes from multi-aspect reviews.
\newblock In {\em Data Mining (ICDM), 2012 IEEE 12th International Conference
  on}, pages 1020--1025. IEEE, 2012.

\bibitem{mcauley2015inferring}
J.~McAuley, R.~Pandey, and J.~Leskovec.
\newblock Inferring networks of substitutable and complementary products.
\newblock In {\em Proceedings of the 21th ACM SIGKDD International Conference
  on Knowledge Discovery and Data Mining}, pages 785--794. ACM, 2015.

\bibitem{mikolov2010recurrent}
T.~Mikolov, M.~Karafi{\'a}t, L.~Burget, J.~Cernock{\`y}, and S.~Khudanpur.
\newblock Recurrent neural network based language model.
\newblock In {\em INTERSPEECH}, pages 1045--1048, 2010.

\bibitem{mikolov2013distributed}
T.~Mikolov, I.~Sutskever, K.~Chen, G.~S. Corrado, and J.~Dean.
\newblock Distributed representations of words and phrases and their
  compositionality.
\newblock In {\em Advances in neural information processing systems}, pages
  3111--3119, 2013.

\bibitem{nair2010rectified}
V.~Nair and G.~E. Hinton.
\newblock Rectified linear units improve restricted boltzmann machines.
\newblock In {\em Proceedings of the 27th International Conference on Machine
  Learning (ICML-10)}, pages 807--814, 2010.

\bibitem{pan2008one}
R.~Pan, Y.~Zhou, B.~Cao, N.~N. Liu, R.~Lukose, M.~Scholz, and Q.~Yang.
\newblock One-class collaborative filtering.
\newblock In {\em Data Mining, 2008. ICDM'08. Eighth IEEE International
  Conference on}, pages 502--511. IEEE, 2008.

\bibitem{rendle2012factorization}
S.~Rendle.
\newblock Factorization machines with libfm.
\newblock {\em ACM Transactions on Intelligent Systems and Technology (TIST)},
  3(3):57, 2012.

\bibitem{DBLP:conf/nips/SalakhutdinovM07}
R.~Salakhutdinov and A.~Mnih.
\newblock Probabilistic matrix factorization.
\newblock In {\em {NIPS}}, pages 1257--1264. Curran Associates, Inc., 2007.

\bibitem{salakhutdinov2007restricted}
R.~Salakhutdinov, A.~Mnih, and G.~Hinton.
\newblock Restricted boltzmann machines for collaborative filtering.
\newblock In {\em Proceedings of the 24th international conference on Machine
  learning}, pages 791--798. ACM, 2007.

\bibitem{schein2002methods}
A.~I. Schein, A.~Popescul, L.~H. Ungar, and D.~M. Pennock.
\newblock Methods and metrics for cold-start recommendations.
\newblock In {\em Proceedings of the 25th annual international ACM SIGIR
  conference on Research and development in information retrieval}, pages
  253--260. ACM, 2002.

\bibitem{srivastava2014dropout}
N.~Srivastava, G.~Hinton, A.~Krizhevsky, I.~Sutskever, and R.~Salakhutdinov.
\newblock Dropout: A simple way to prevent neural networks from overfitting.
\newblock {\em The Journal of Machine Learning Research}, 15(1):1929--1958,
  2014.

\bibitem{2016arXiv160502688short}
{Theano Development Team}.
\newblock {Theano: A {Python} framework for fast computation of mathematical
  expressions}.
\newblock {\em arXiv e-prints}, abs/1605.02688, May 2016.

\bibitem{tieleman2012lecture}
T.~Tieleman and G.~Hinton.
\newblock Lecture 6.5-rmsprop: Divide the gradient by a running average of its
  recent magnitude.
\newblock {\em COURSERA: Neural Networks for Machine Learning}, 4:2, 2012.

\bibitem{van2013deep}
A.~Van~den Oord, S.~Dieleman, and B.~Schrauwen.
\newblock Deep content-based music recommendation.
\newblock In {\em Advances in Neural Information Processing Systems}, pages
  2643--2651, 2013.

\bibitem{wallach2006topic}
H.~M. Wallach.
\newblock Topic modeling: beyond bag-of-words.
\newblock In {\em Proceedings of the 23rd international conference on Machine
  learning}, pages 977--984. ACM, 2006.

\bibitem{wang2011collaborative}
C.~Wang and D.~M. Blei.
\newblock Collaborative topic modeling for recommending scientific articles.
\newblock In {\em Proceedings of the 17th ACM SIGKDD international conference
  on Knowledge discovery and data mining}, pages 448--456. ACM, 2011.

\bibitem{wang2010latent}
H.~Wang, Y.~Lu, and C.~Zhai.
\newblock Latent aspect rating analysis on review text data: a rating
  regression approach.
\newblock In {\em Proceedings of the 16th ACM SIGKDD international conference
  on Knowledge discovery and data mining}, pages 783--792. ACM, 2010.

\bibitem{wang2015collaborative}
H.~Wang, N.~Wang, and D.-Y. Yeung.
\newblock Collaborative deep learning for recommender systems.
\newblock In {\em Proceedings of the 21th ACM SIGKDD International Conference
  on Knowledge Discovery and Data Mining}, pages 1235--1244. ACM, 2015.

\bibitem{wang2014improving}
X.~Wang and Y.~Wang.
\newblock Improving content-based and hybrid music recommendation using deep
  learning.
\newblock In {\em Proceedings of the ACM International Conference on
  Multimedia}, pages 627--636. ACM, 2014.

\bibitem{wucollaborative}
Y.~Wu, C.~DuBois, A.~X. Zheng, and M.~Ester.
\newblock Collaborative denoising auto-encoders for top-n recommender systems.

\bibitem{WuE15}
Y.~Wu and M.~Ester.
\newblock {FLAME:} {A} probabilistic model combining aspect based opinion
  mining and collaborative filtering.
\newblock In {\em {WSDM}}, pages 199--208. {ACM}, 2015.

\end{thebibliography}
\end{document}